\PassOptionsToPackage{table}{xcolor}
\documentclass[twocolumn]{fairmeta}

\usepackage{multirow}
\usepackage{arydshln}
\usepackage{subcaption}
\usepackage{stfloats}
\usepackage{placeins}
\usepackage{pifont}
\usepackage{listings}
\usepackage{xspace}
\usepackage{amsfonts}

\makeatletter
\DeclareRobustCommand\onedot{\futurelet\@let@token\@onedot}
\def\@onedot{\ifx\@let@token.\else.\null\fi\xspace}

\definecolor{redorange}{HTML}{F26035}
\definecolor{cerulean}{HTML}{00A2E3}
\definecolor{olive}{HTML}{3C8031}
\definecolor{violet}{HTML}{99479B}

\usepackage{float}
\newfloat{prompt}{thp}{lop}
\floatname{prompt}{Prompt}

\newcommand{\ourbenchmark}{\textsc{MVP}}

\usepackage{tabularx}
\usepackage{placeins}
\usepackage{graphicx}
\usepackage{subcaption}
\usepackage{float}
\usepackage{array}

\usepackage{soul}
\usepackage[table]{xcolor}
\usepackage{makecell}
\usepackage{multirow}
\usepackage{tcolorbox}

\setul{0.9ex}{0.15ex}  
\setulcolor{white}      
\usepackage{makecell}

\usepackage{booktabs}
\usepackage[switch]{lineno}

\definecolor{cvprblue}{rgb}{0.21,0.49,0.74}

\title{A Shortcut-aware Video-QA Benchmark for Physical Understanding via Minimal Video Pairs}

\author[1,2,3,*]{Benno Krojer}
\author[1]{Mojtaba Komeili}
\author[1]{Candace Ross}
\author[1]{Quentin Garrido}
\author[1]{Koustuv Sinha}
\author[1]{Nicolas Ballas}
\author[1]{Mahmoud Assran}

\affiliation[1]{FAIR at Meta}
\affiliation[2]{Mila}
\affiliation[3]{McGill University}

\contribution[*]{Work done during internship}

\abstract{
Existing benchmarks for assessing the spatio-temporal understanding and reasoning abilities of video language models are susceptible to score inflation due to the presence of shortcut solutions based on superficial visual or textual cues.
This paper mitigates the challenges in accurately assessing model performance by introducing the Minimal Video Pairs (\textbf{\ourbenchmark{}}) benchmark, a simple shortcut-aware video QA benchmark for assessing the physical understanding of video language models.
The benchmark is comprised of 55K high-quality multiple-choice video QA examples focusing on physical world understanding.
Examples are curated from nine video data sources, spanning first-person egocentric and exocentric videos, robotic interaction data, and cognitive science intuitive physics benchmarks.
To mitigate shortcut solutions that rely on superficial visual or textual cues and biases, each sample in \ourbenchmark{} has a minimal-change pair --- a visually similar video accompanied by an identical question but an opposing answer.
To answer a question correctly, a model must provide correct answers for both examples in the minimal-change pair; as such, models that solely rely on visual or textual biases would achieve below random performance.
Human performance on \ourbenchmark{} is 92.9\%, while the best open-source state-of-the-art video-language model achieves 40.2\% compared to random performance at 25\%.\looseness-1
}

\date{\today}
\correspondence{Benno Krojer at \email{benno.krojer@mila.quebec}}
\begin{document}

\maketitle

\begin{figure*}[ht]
    \centering
    \includegraphics[width=\textwidth]{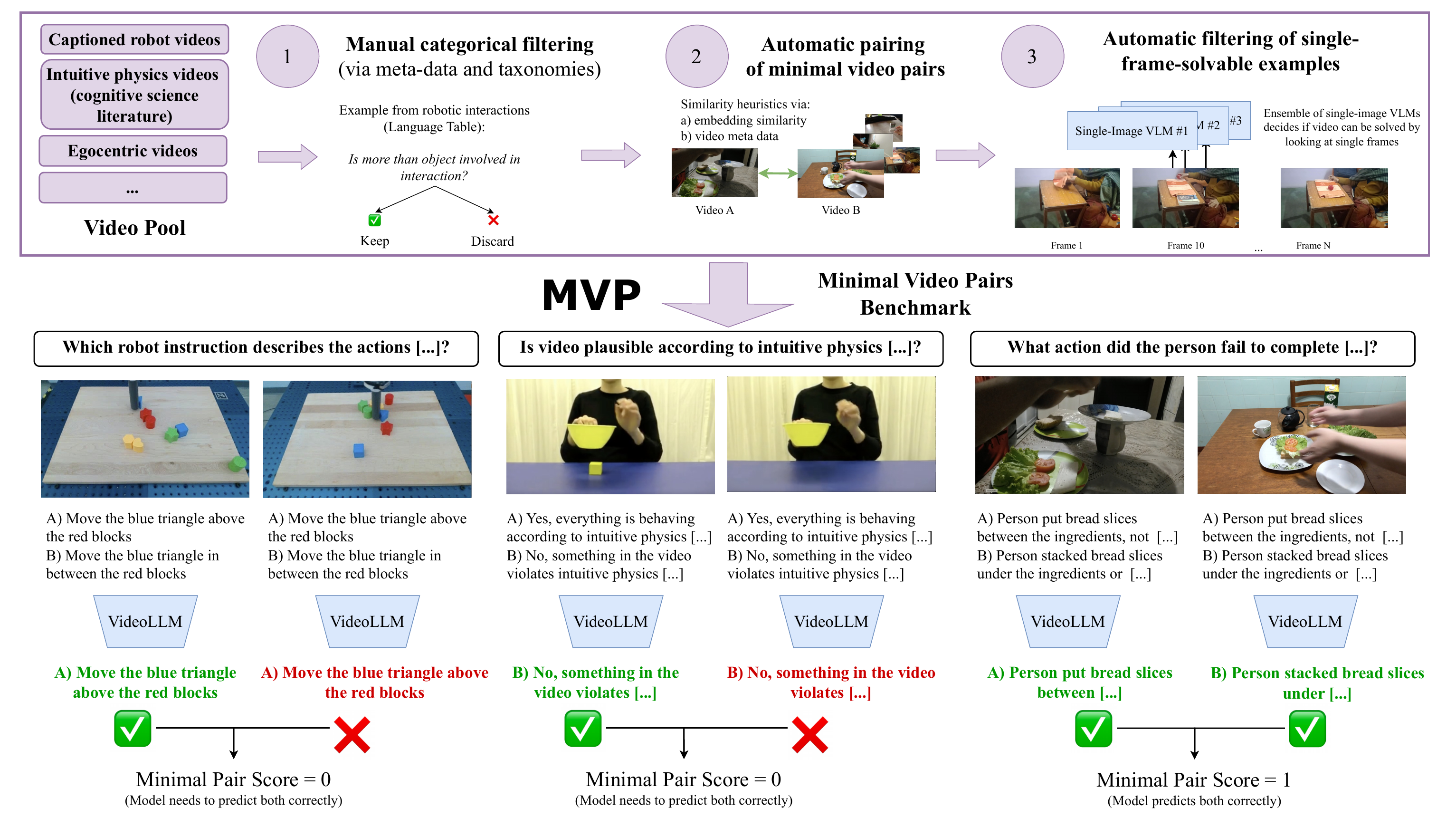}
    \label{fig:main}
        \captionof{figure}{Illustrating \ourbenchmark{} with its curation steps (top) and examples of our Minimal Pair Scoring (bottom). \looseness-1}
\end{figure*}

\section{Introduction}
\label{sec:intro}

{\it Moravec's paradox} highlights a counterintuitive phenomenon:~high-level reasoning tasks, often perceived as complex, are typically easier for AI agents to solve than sensorimotor and perception tasks, which are seemingly effortless for humans~\citep{moravec1988mind}.

Recently, large vision-language models have emerged as a promising paradigm for enabling perception capabilities in AI agents, demonstrating impressive progress on question-answering tasks across various domains including movies, documents, charts, and sciences~\citep{alayrac2022flamingo, team2024gemini,dubey2024llama, wang2024tarsier}. This progress raises a natural question: do these models possess the spatiotemporal understanding and reasoning abilities essential for an agent to interact within the physical world, or do they buttress Moravec's paradox?

Various visual QA datasets have been proposed by the community to assess the spatiotemporal understanding of video-language models~\citep{tapaswi2016movieqa,maharaj2017dataset,li2024mvbench,patraucean2023perception,zhang2023movqa,xie2025funqa,wang2023paxion,Yi2020CLEVRER}; one of the most popular, MVBench~\citep{li2024mvbench}, combines 11 video datasets into a single video QA benchmark.

While recent state-of-the-art video-language models obtain performance far superior to a random baseline on these benchmarks~\citep{wang2024tarsier, shen2024longvu, li2024llava}, our investigation reveals that existing models can achieve strong performance on these benchmarks by relying on superficial visual or textual cues or biases.
This is validated using simple baselines that discard the visual input or temporal aspect, yet achieve non-trivial performance.
Similarly, concurrent work~\citep{cores2024tvbench} shows that some of these tasks~\citep{li2024mvbench} fail to accurately measure the temporal understanding of a model.

In this work, we take inspiration from works in natural language processing~\citep{levesque2012winograd,sakaguchi2021winogrande} and image processing~\citep{thrush2022winoground,yuksekgonul2022and} addressing visual and textual biases in evaluation, and introduce \ourbenchmark{}, a video QA benchmark containing {\underline m}inimal-change {\underline v}ideo {\underline p}airs.
Specifically, each video-question-answer sample in the benchmark is accompanied by a visually similar video possessing an identical question but an opposing answer (Figure~\ref{fig:main}).
To answer a question correctly, a model must also provide the correct answer for its minimal-change pair while \textit{processing them independently}.
Many types of shortcut solutions are penalized under the minimal-pair scoring framework as a model relying on superficial visual or textual cues or biases would incorrectly output the same answer for both the samples in the pair.

While recent work created small sets of minimal-change video pairs for course-grained temporal reasoning~\citep{zhang2024vinoground,liu-etal-2024-tempcompass}, our key insight is that these pairs can be efficiently mined from existing video sources to test for several model capabilities through an automated process relying on visual embeddings and video meta-data.
We propose an automatic process to find minimally different video pairs with limited human intervention, and then build these into a video-question-answer tuple with identical questions and opposing answers, enabling the scaling of the benchmark to a broad set of videos spanning diverse situations.
We further process the mined samples using a model ensemble to filter out single-frame solvable examples --- questions that can be answered using any single randomly sampled frame from the video --- to encourage a stronger focus on video understanding.
We build \ourbenchmark{} by running our process on nine video sources spanning intuitive physics understanding, spatiotemporal reasoning, action anticipation, and robotic manipulation, leading to a total of $54,828$ multiple-choice video QA examples with minimal-change pairs; i.e. $27,414$ minimal-change pairs.

Next, we assess recent proprietary and open-source state-of-the-art video-language models using \ourbenchmark{}.
Specifically, we evaluate 2 closed-source models (GPT4-o~\citep{achiam2023gpt} and Gemini-1.5 Pro~\citep{team2024gemini}), and 7 open-source video-language models: LLaVA-OneVision ~\cite{li2024llava}, VideoChat2~\cite{li2024mvbench}, Mini-CPM~\citep{yao2024minicpm}, Qwen2-VL~\citep{bai2023qwen}, Tarsier~\citep{wang2024tarsier}, LongVu~\citep{shen2024longvu}, InternVL-2.5~\citep{chen2024expanding}.
We find that even proprietary models are only slightly above random and that the best accuracy achieved across models is only $40.2\%$ , in stark contrast to human baseline performance at 92.9\% accuracy.
These findings suggest that video-language models may still struggle with seemingly simple physical reasoning tasks, despite achieving relatively high accuracy on standard spatio-temporal reasoning benchmarks.\\[-1ex]

\noindent
In short, we make the following contributions:
\begin{enumerate}
    \item Analyze potential shortcut solutions on all 11 datasets in the popular MVBench~\citep{li2024mvbench} benchmark suite, using simple baselines consisting of language-only models, single-frame/image models, and Socratic LLMs.
    \item Introduce \ourbenchmark{}, a video QA benchmark for physical world understanding comprising minimally different videos --- the largest of its kind by an order of magnitude with $\sim$55K examples.
    \item Benchmark closed-source and open-source state-of-the-art models and identify a gap in physical world understanding; human performance on \ourbenchmark{} is around 92.9\%, while even GPT4-o and Gemini achieve around 30\% compared to random performance at 25\%.
\end{enumerate}

\begin{figure}
    \centering
    \vspace{1ex}
    \includegraphics[width=\linewidth]{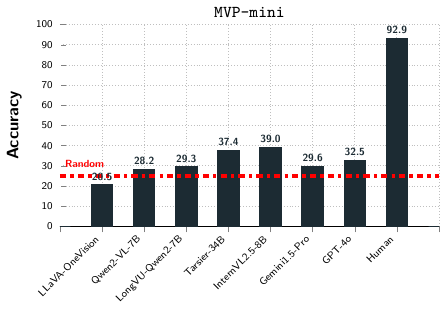}
    \caption{ Performance of the strongest evaluated \mbox{VideoLLMs} on \ourbenchmark{} (mini-version), compared to human performance.}
    \label{fig:main}
\end{figure}
\section{Robustness Analysis of MVBench}
\label{sec:shortcuts}

We begin by examining robustness of existing video QA benchmarks to shortcut solutions based on visual or textual cues or biases.
Specifically, our analysis focuses on CLEVRER~\citep{johnson2017clevr}, Perception Test~\citep{patraucean2023perception}, STAR~\citep{wu2021star_situated_reasoning}, PAXION~\citep{wang2023paxion}, Moments in Time V1~\citep{mit}, FunQA~\citep{xie2025funqa}, Charades-STA~\citep{charades_sta}, MoVQA~\citep{movqa}, NTU RGB+D~\citep{ntu_rgbd}, VLN-CE~\citep{vln_ce} and TVQA~\citep{lei-etal-2018-tvqa}, which are all included in the widely adopted MVBench~\citep{li2024mvbench} benchmark suite.

\setlength{\arrayrulewidth}{0.4pt}

\begin{table*}[h]
\centering
\setlength{\tabcolsep}{1pt} 
\renewcommand{\arraystretch}{1.1}
\begin{minipage}{\textwidth}
{\footnotesize
\begin{tabular}{c|c|c|c|c|c|c|c|c|c|c|c|c|c|c|c|c|c|c|c|c|c}
    \hline
    \bf Task & \cellcolor{gray!20} Avg & AA & AC & AL & AP & AS & CI & CO & EN & ER & FA & FP & MA & MC & MD & OE & OI & OS & ST & SC & UA \\
    \hline
    \it Random Chance & \cellcolor{gray!20} 0.30 & 0.50 & 0.33 & 0.25 & 0.25 & 0.25 & 0.31 & 0.33 & 0.25 & 0.20 & 0.25 & 0.25 & 0.33 & 0.25 & 0.25 & 0.50 & 0.25 & 0.33 & 0.25 & 0.33 & 0.25 \\
    \hline
    GPT-4V \textsuperscript{\dag}  & \cellcolor{gray!20} 0.44 & 0.72 & 0.39 & 0.41 & \textbf{0.64} & 0.56 & 0.11 & 0.52 & 0.31 & \textbf{0.59} & 0.47 & 0.48 & 0.23 & 0.12 & 0.12 & 0.19 & 0.59 & 0.30 & 0.84 & 0.45 & 0.74 \\
    VideoChat2 (Mistral) &  \cellcolor{gray!20} \textbf{0.61} & 0.86 & 0.37 & \textbf{0.44} & 0.55 & \textbf{0.76} & 0.72 & 0.49 & 0.36 & 0.40 & 0.50 & 0.64 & \textbf{0.88} & \textbf{0.69} & \textbf{0.49} & \textbf{0.87} & 0.75 & \textbf{0.41} & \textbf{0.85} & \textbf{0.50} & 0.62 \\
    \hline
    \multicolumn{22}{c}{\textit{\textbf{Language only:} Model considers question and answer choices, without access to the video.}} \\
    Llama 3-8B & \cellcolor{gray!20}0.36 & 0.63 & \ul{0.38} & 0.27 & 0.25 & 0.28 & 0.35 & 0.43 & 0.29 & 0.43 & 0.29 & 0.29 & 0.38 & 0.27 & 0.21 & 0.46 & 0.29 & 0.36 & 0.52 & 0.40 & 0.52 \\
    Llama 3-70B & \cellcolor{gray!20}0.38 & \ul{0.78} & \ul{0.39} & 0.32 & 0.26 & 0.26 & 0.43 & \ul{0.47} & 0.28 & 0.46 & 0.26 & 0.27 & 0.41 & 0.29 & 0.20 & 0.48 & 0.29 & 0.32 & 0.48 & \ul{0.45} & \ul{0.58} \\
    \hline
    \multicolumn{22}{c}{\textit{\textbf{Video only:} Model considers video and answer choices only, without access to the question.}} \\
    VideoChat2 (Mistral) & \cellcolor{gray!20} 0.50 & \textbf{\ul{0.88}} & \textbf{\ul{0.42}} & 0.25 & 0.49 & \ul{0.68} & \textbf{\ul{0.74}} & 0.44 & 0.28 & 0.39 & \textbf{\ul{0.53}} & \textbf{\ul{0.65}} & 0.47 & 0.29 & 0.26 & 0.53 & \textbf{\ul{0.75}} & 0.34 & \ul{0.81} & 0.32 & \ul{0.55} \\
    \hline
    \multicolumn{22}{c}{\textit{\textbf{Single-Frame only:} Model considers question, answer choices and a single key frame, without access to the full video.}} \\
    Idefics3 & \cellcolor{gray!20}0.47 & 0.72 & \ul{0.37} & 0.31 & 0.52 & 0.48 & 0.42 & \ul{0.54} & 0.31 & \ul{0.48} & 0.40 & 0.44 & 0.55 & 0.42 & 0.34 & 0.49 & 0.50 & 0.37 & \ul{0.73} & \ul{0.48} & \ul{0.60} \\
    Qwen2-VL & \cellcolor{gray!20}0.51 & \ul{0.87} & \ul{0.37} & 0.31 & \ul{0.55} &0.54 &0.57 & \textbf{\ul{0.59}} & \textbf{\ul{0.40}} &0.45 &\ul{0.46} &\ul{0.53} &0.6 &0.43 &0.37 &0.53 &0.54 &\ul{0.39} &\ul{0.74} &0.42 &\textbf{\ul{0.68}} \\
    \hline
    \multicolumn{22}{c}{\textit{\textbf{Simple Socratic LLM:} Model considers the question, answer choices and a short generic description of the video.}} \\
    \hline
    Llama 3-8B & \cellcolor{gray!20}0.44 & 0.56 & \ul{0.38} & 0.28 & 0.49 & 0.57 & 0.35 & \ul{0.53} & 0.29 & 0.42 & 0.30 & 0.35 & 0.56 & 0.42 & 0.32 & 0.50 & 0.56 & 0.35 & 0.68 & 0.44 & \ul{0.56} \\
    Llama 3-70B & \cellcolor{gray!20}0.46 & 0.67 & 0.32 & 0.35 & 0.40 & 0.55 & 0.38 & \ul{0.55} & 0.24 & 0.45 & 0.36 & 0.41 & 0.56 & 0.46 & 0.32 & 0.57 & 0.62 & 0.35 & 0.70 & 0.39 & 0.54 \\
    \hline
\end{tabular}
}
\caption{\textbf{Shortcut Analysis on the 20 MVBench tasks from 11 datasets}:
Optimal performance on these spatio-temporal reasoning benchmarks is frequently achieved by models relying on visual or textual biases (Single-Frame only, Video only, Simple Socratic LLM). \textsuperscript{\dag}: GPT-4V accuracy from \citep{li2024mvbench}.
\scriptsize{
Tasks: AA (Action Antonym), AC (Action Count), AL (Action Localization), AP (Action Prediction), AS (Action Sequence), CI (Counterfactual Inference), CO (Character Order), EN (Egocentric Navigation), ER (Episodic Reasoning), FA (Fine-grained Action), FP (Fine-grained Pose), MA (Moving Attribute), MC (Moving Count), MD (Moving Direction), OE (Object Existence), OI (Object Interaction), OS (Object Shuffle), ST (Scene Transition), SC (State Change), UA (Unexpected Action).}
}
\label{tab:shortcut_analysis}
\end{minipage}
\end{table*}

\noindent\textbf{Empirical Setup.}
MVBench is comprised of 20 tasks from 11 datasets, collected in a multiple-choice video QA format, where a model is required to choose an answer $a_i$ from a tuple of question, video, and answer candidates \((q, v, [\text{a}_1, \text{a}_2, ..])\).
Following standard practice~\citep{goyal2017making}, we study robustness to shortcuts by perturbing the task inputs, e.g., requiring the model to select an answer candidate without seeing the video or perhaps without reading the question, and compare to the accuracy achieved by a video LLM without perturbing the task inputs. 
We study 4 types of shortcut solutions by evaluating language-only models, video-only models, single-frame models, and simple Socratic LLMs.
Results are reported in Table~\ref{tab:shortcut_analysis} using the original skill taxonomy outlined in MVBench.

\noindent\textbf{Language only.}
Language-only models do not observe the video, and therefore select an answer candidate by only considering the textual inputs $q$ and the answer candidates \([\text{a}_1, \text{a}_2, ..]\).
We leverage the Llama3-8B and Llama3-70B models due to their competitive performances~\citep{dubey2024llama}.
In Table~\ref{tab:shortcut_analysis}, we find that a Llama3-8B outperforms a random baseline by 6\%, and a larger Llama3-70B outperforms a random baseline by 8\%, suggesting that only a small subset of examples can be solved without considering the video input.
However, digging into the individual datasets and sub-tasks in Table~\ref{tab:shortcut_analysis} reveals strong language-only performance on Action Antonym, where LLaMA3-70 achieves 78\% compared to a random baseline at 50\%.
Upon closer inspection of the original dataset, we observe that many questions can be correctly selected by choosing the answer candidate with the highest marginal likelihood.
For instance, given an example with answer candidates ``book falling like a rock`` versus ``book rising like a rock,'' an LLM, just like a human, can rely on its language bias to infer that the former is probably the correct description without observing the video.

\noindent\textbf{Video only.}
Video-only models do not observe the question, and therefore select an answer candidate by only considering the video input $v$ and the answer candidates $[\text{a}_1, \text{a}_2, ..]$.
Table~\ref{tab:shortcut_analysis} shows that a video LLM (VideoChat2-Mistral) can solve most sub-tasks without access to the question, reaching 50\% overall accuracy; by comparison the same model achieves an accuracy of 61\% when given the question in addition to the video, while a random baseline is at 30\%.
These findings indicate that the answer candidates for each question $[\text{a}_1, \text{a}_2, ..]$ are not sufficiently task-specific, as the model is able to discard the incorrect answers without knowing question.

This trend is particularly interesting on the  counterfactual inference sub-task, where the counterfactual scenario such as ``What happens if the cube is removed?'' can only be known from the question.
Manual inspection reveals that the correct answer in this task (based on CLEVRER ~\citep{Yi2020CLEVRER}) often occurs regardless of the counterfactual scenario, e.g., the two objects in question will collide regardless of the causal intervention.

\noindent\textbf{Single-frame only.}
Single-frame models do not observe the entire video, but rather are provided only a single frame $f_i \in v$ form the video.
These models must therefore select an answer candidate by considering the frame $f_i$, the textual inputs $q$, and the answer candidates $[\text{a}_1, \text{a}_2, ..]$.
We take $f_i$ to be the center frame from the video and leverage Idefics3-8B~\citep{laurenccon2024building} and Qwen2-VL-7B~\citep{wang2024qwen2} for the single-frame baselines.
In Table~\ref{tab:shortcut_analysis}, Idefics3-8B achieves an overall accuracy of 47\% and Qwen2-VL-7B achieves an overall 51\% accuracy, which is comparable to the performance of full-fledged VideoLLMs.
Moreover, on Action Antonym, Action Prediction, Character Order, Egocentric Navigation, Episodic Reasoning, Fine-grained Action, State Transition, and Unexpected Action, the single-frame models are on par with (or even exceed) the performance of the VideoLLMs.
Concurrent work~\citep{cores2024tvbench} also studies the related bag-of-frame bias by shuffling the video frames.

\noindent\textbf{Simple Socratic LLM.}
A Simple Socratic LLM~\citep{zhang2023simple, zeng2023socratic} replaces the video input $v$ with a short caption $c_v$ that can only convey a low-bandwidth description of the video.
In practice, $c_v$ is 1 or 2 sentence-long caption generated by a separate VideoLLM~\citep{zhang2024internlm} in a task-independent manner.
The Socratic LLMs therefore select an answer candidate by only considering the low-bandwidth caption $c_v$, the question $q$, and the answer candidates $[\text{a}_1, \text{a}_2, ..]$.
Following the text-only baselines, we use Llama3-8B and 70B.
The performance of the Simple Socratic LLMs in Table~\ref{tab:shortcut_analysis} is significantly above random, with 44\% for the LLaMA3-8B and 47\% for the LLaMA-70B, suggesting that many sub-tasks (e.g. Character Order, Episodic Reasoning, Scene Transition) do not require fine-grained scene understanding.

\noindent\textbf{Summary.}
The shortcut analysis reveals that existing models can often achieve strong performance on spatio-temporal reasoning benchmarks by relying on language cues (\textit{Language only} shortcut) or visual cues, (\textit{Video only} shortcut), and may not need to perform temporal reasoning (\textit{Single-Frame only} shortcut), or possess fine-grained visual features (\textit{Simplified Socratic LLM}).
\definecolor{darkerGreen}{rgb}{0.0, 0.7, 0.0} 
\newcommand{\darkerGreen}{\textcolor{darkerGreen}}
\begin{table*}[ht]
\vskip 0.15in
\begin{center}
\begin{footnotesize}
\begin{sc}
\setlength{\tabcolsep}{3.4pt} 
\begin{tabular}{lcccccccccc}
\toprule
& \multicolumn{5}{c}{Domains of VideoQA-Examples} & & & & \\
\cmidrule(lr){2-6}
& Total & Natural  & Intuitive & Robotics & Synthetic & Minimially & Procedural  &  \\
& & Videos & Physics & & Videos & Diff. & Single-Frame & \\
Benchmark & & & & & & Videos & Bias Filtering & Format \\
\midrule
CLEVRER & \darkerGreen{76.3K} & \textcolor{red}{0K} & \textcolor{gray}{21.4K} & \textcolor{red}{0K} & \darkerGreen{76.3K} &  \textcolor{red}{\texttimes} & \textcolor{red}{\texttimes} & MC-QA \\
Perception Test & \darkerGreen{11.5K} & \darkerGreen{11.5K} & \textcolor{red}{0-0.2K} & \textcolor{red}{0K} & \textcolor{red}{0K} &  \textcolor{red}{\texttimes} & \textcolor{red}{\texttimes} & MC-QA \\
MVBench & 4K & 2.8K & \textcolor{red}{0K} & \textcolor{red}{0.2K} & 1.2K &  \textcolor{red}{\texttimes} & \textcolor{red}{\texttimes} & MC-QA \\
TVBench & 2.5K & 1.9K & \textcolor{red}{0K} & \textcolor{red}{0.2K} & \textcolor{red}{0.6K} & \textcolor{red}{\texttimes} & \textcolor{red}{\texttimes} & MC-QA \\
Vinoground & \textcolor{red}{1K} & 1K & \textcolor{red}{0K} & \textcolor{red}{0K} & \textcolor{red}{0K} & \darkerGreen\checkmark & \textcolor{red}{\texttimes} & Group-Score \\
TempCompass & \textcolor{red}{0.5K} & \textcolor{red}{0.5K} & \textcolor{red}{0K} & \textcolor{red}{0K} & \textcolor{red}{0K} & \darkerGreen\checkmark & \textcolor{red}{\texttimes} & Group-Score \\
\midrule
\ourbenchmark{} & \darkerGreen{54.8K} & \darkerGreen{22.3K} & \darkerGreen{9.9K} & \darkerGreen{25.8K} & \darkerGreen{32.6K} & \darkerGreen\checkmark & \darkerGreen\checkmark & Pair MC-QA \\
\bottomrule

\end{tabular}
\caption{\textbf{We compare with recent benchmarks that focus on similar skills}. Note that some videos may fall within several categories (e.g., synthetic intuitive physics videos). \ourbenchmark{} contains minimally different videos at a much larger scale and across more diverse domains.
From these benchmarks, \ourbenchmark{} is the first to procedurally filter out examples due to single-frame bias.
\textsc{Group-Score} = Present one video + two captions, and two videos + one caption.
CLEVRER's intuitive physics entry is grayed as it only covers a narrow subset of intuitive physics concepts, largely based on collisions.
}
\label{tab:benchmark_comparison}

\end{sc}
\end{footnotesize}
\end{center}
\vskip -0.1in
\end{table*}
\begin{table*}[t]
    \centering
    \renewcommand{\arraystretch}{1.2}
    \setlength{\tabcolsep}{4pt}
    \footnotesize
    \begin{tabular}{p{4cm} p{5cm} p{7cm}}
        \toprule
        \textbf{Benchmark Category} & \textbf{Sources (\# paired video-QA examples)} & \textbf{Example} \\
        \midrule
        Fine-grained human-object interactions & 
        Perception Test (3.5K), Something Something v2 (3.6K) & \textbf{Q:} \textit{What stops the motion of the object placed on the slanted plane after being released [...]?} \textbf{A)} Person or collision with another object \textbf{B)} High friction with surface \\ 
        \midrule
        Fine-grained robot-object interactions & 
        Language Table (12.9K) & \textbf{Q:} \textit{Which robot instruction best describes the actions in the video?} \textbf{A)} Move the green blocks in a vertical line below blue cube \textbf{B)} Move the green blocks and blue cube in a vertical line \\ 
        \midrule
        Intuitive physics and collisions & 
        IntPhys (0.2K), InfLevel (2.6K), GRASP (2.0K), CLEVRER (1.2K) & \textbf{Q:} \textit{Is this video physically plausible/possible according to your understanding of e.g. object permanence, gravity, [...]} \textbf{A)} Yes, everything is behaving according to human intuitive physics understanding \textbf{B)} No, something in the video is off/strange or violates [...] \\ 
        \midrule
        Coarse-grained temporal reasoning & 
        STAR (1.0K), Vinoground (0.5K) & \textbf{Q:} \textit{What is the best caption for this video?} \textbf{A)} The kayak flips over from facing upwards towards facing downwards \textbf{B)} The kayak flips over from facing downwards towards facing upwards \\ 
        \bottomrule
    \end{tabular}
    \caption{\textbf{Overview of \ourbenchmark{}}. Each answer option A/B is correct for only one video in the minimal-change pair, while acting as a hard negative for the other video. Note that we show the number of \textit{paired} video-QA examples, thus the number of videos in our data is twice that amount.}
    \label{tab:benchmark_overview}
\end{table*}

\section{Testing  Physical World Understanding via Minimal Change Pairs}
\label{sec:benchmark}
In this section we discuss the construction of \ourbenchmark{} to mitigate shortcut solutions based on visual and textual biases.
\ourbenchmark{} is comprised of $54,828$ video QA examples covering various aspects of physical world understanding, including spatial reasoning, temporal understanding, human-object interaction, memory, counterfactuals, anticipation, and intuitive physics.

\noindent\textbf{Task formulation.}
To improve robustness to the various shortcut solutions described in the previous section, we adopt a minimal-change pair approach~\citep{levesque2012winograd,sakaguchi2021winogrande}. 
An example in \ourbenchmark{} consists of two video QA pairs \((q_1, v_1, [\text{a}_1, \text{a}_2])\) and \((q_2, v_2, [\text{a}_1, \text{a}_2])\) containing identical questions $q_1 = q_2$, visually similar videos $v_1 \sim v_2$, and two mutually exclusive (i.e., contradicting) answer candidates $a_1$ and $a_2$.

\noindent\textbf{Minimal-change Pair Scoring.}
A model relying on superficial visual or textual cues or biases to solve a task will tend to produce the same output for each sample in the minimal-change pair.
Thus, to penalize models for latching onto shortcuts, we only provide a positive score if the correct answer is produced for both minimal-change samples; the model receives each example \((q, v_1, [\text{a}_1, \text{a}_2])\) and \((q, v_2, [\text{a}_1, \text{a}_2])\) in isolation.
Following a multiple choice QA framework, the model has to output a single answer letter (A or B) via task-specific prompts. In this setup, a random baseline achieves an accuracy of 25\%.

\noindent\textbf{Question Taxonomy.}
We wish to understand whether video LLMs possess the spatio-temporal understanding and reasoning abilities essential for an agent to interact within the physical world.
As such we consider a coarse-grained taxonomy of question categories encompassing:
\begin{itemize}
    \item Fine-grained human-object interactions,
    \item Fine-grained robot-object interactions,
    \item Intuitive Physics understanding,
    \item Coarse-grained temporal understanding.
\end{itemize}

We intentionally construct samples that are not overly reliant on cultural knowledge~\citep{Rawal2024CinePileAL,he2024mmworld,Li2024VideoVistaAV} (e.g., movies) or specific domain knowledge~\citep{Tang_2019_CVPR} (e.g., detailed recipes) --- tasks where language bias could contribute to the general performance.

We first manually filter videos from the sources described in Table~\ref{tab:benchmark_overview} based on manual inspection (cf.~\Cref{app:manual_filter}), then convert them into a question-answer format based on the associated meta-data (the textual captions for Language Table, the class labels for Something-Something-v2, QA annotations for PT, Vinoground, STAR, and CLEVERER, and the concept labels for IntPhys, InfLevel, and GRASP), yielding a starting set of 548K video QA examples.

\noindent\textbf{Minimal-change Pair Mining.}
Next we procedurally identify minimal-change pairs from the 548K video QA examples produced from the previous stage.
We note that 16\% of the videos in our final benchmark ($\sim$ 8.8K examples) already possess explicit minimal visual pairs.
For the remaining 84\% of the videos, we leverage the following procedure to construct visual minimal-change pairs.
In this process, we search for samples that have visually similar videos, identical questions (based on semantic matching), and contradictory answers.
To then determine whether two videos with the same question are suitable minimal pairs, we use a) \textbf{symbolic and neural rules to determine video similarity} and b) \textbf{entailment detection}~\citep{bowman-etal-2015-large,dagan2013recognizing} between the correct answers of each video.
Whether we rely more on symbolic or or neural rules of similarity depends on the data source:
If a dataset has rich annotations (positions or attributes of objects) or structured captions (such as CLEVRER or Something Something-v2), we use hand-crafted rules and the NLP toolkit spacy~\citep{spacy2} to narrow down the candidate pool of minimal pairs.
This step would match videos with a large intersection of objects or attributes mentioned in the annotation/caption, leading to highly similar videos (e.g., the same objects appearing in both videos).
Once we have narrowed down the pool of candidate pairs, in the final step we rank video pairs by their cosine similarity in the ViCLIP~\citep{wang2023internvid} video embedding space.
We then select the top-ranked minimal video pairs such that each question or skill-type is sufficiently represented.
At the same time, we ensure that the correct answers for samples in a minimal-change pair are sufficiently different, as the correct answer of one element in the pair must be a truly negative (\textit{negative}) answer candidate for the other element, and vice versa:
To avoid cases where both answers could be true at the same time (e.g., synonyms or more subtle cases) we define a set of textual rules to detect entailment for a subset of datasets.
To illustrate this, in the \textit{Fine-grained Robot-object interactions} category, our entailment-detection would discard the following pair of answers: A) ``Move the blue cube towards the red heart'' and B) ``Move the blue cube to the left of the red heart'', since A entails B.
After this minimal-pair mining, we are down to 70K QA examples; cf.~\Cref{app:minimal_filtering} for technical details of the minimal pair mining process.

\noindent\textbf{Single-frame Bias Filtering.}
Finally, to address single-frame bias, we remove examples that can be solved without the temporal information in the video; i.e., using only a single frame.
We note that the input frame for this filtering stage should not be selected in a ``smart way,'' since key-frame selection can be regarded as a basic form of temporal reasoning.
In practice, five state-of-the-art multi-modal LLMs (LLama3.2-11B~\citep{dubey2024llama}, Molmo-7B~\citep{deitke2024molmo}, Pixtral-12B~\citep{agrawal2024pixtral}, LLaVA-OneVision-7B~\citep{li2024llava}, Idefics3-8B~\citep{laurenccon2024building}) are prompted to answer the video-QA questions and ``give their best commonsense guess given a single frame sampled from a video.''
If at least 4 out of 5 models in the ensemble predict the correct answer given the same frame, then we flag that frame as \textit{solvable}.
The minimal-change pair is then discarded if 30\% of the frames in both videos are deemed solvable.
This heuristic process removes around 20\% of the samples from the previous stage.

\FloatBarrier
\noindent\textbf{\ourbenchmark{} Statistics.}
We end up with $54,828$ examples in \ourbenchmark{}, grouped into $27,414$ minimal-change video QA pairs.
A breakdown of these examples is shown in \Cref{tab:benchmark_comparison} and \Cref{tab:benchmark_overview} with a reasonably balanced split between natural videos, synthetic videos, robotics videos, and intuitive physics videos.
An average video is 8.8 seconds long, the answer candidates contain an average of 8.1 words, and the datasets contains 2355 unique words in the questions and answers.
Note that the word diversity is much less than MVBench~\citep{li2024mvbench}, which has only 4K examples but twice the number of unique words (4338), reflecting our focus in testing for physical world understanding and not linguistically-diverse tasks with cultural or domain knowledge.
Instead the task difficulty arises from the physical and perceptual aspects of \ourbenchmark{}.

\section{Empirical Results on \ourbenchmark{}}
\label{sec:results}
\begin{table*}[h]
\begin{small}
\centering
\setlength{\tabcolsep}{4pt} 
\begin{tabular}{lccccc}
\toprule
\footnotesize \textbf{Model} & \footnotesize \cellcolor{gray!20}\makecell{\textbf{\ourbenchmark{}} \\ \cellcolor{gray!20}\textbf{(macro-avg)}} & \footnotesize \makecell{\textbf{Fine-grained} \\ \textbf{human-object interactions}} & \footnotesize \makecell{\textbf{Fine-grained} \\ \textbf{robot-object interactions}} & \footnotesize \makecell{\textbf{Intuitive physics} \\ \textbf{and collisions}} & \footnotesize \makecell{\textbf{Coarse-grained} \\ \textbf{temporal reasoning}} \\
\midrule
Random & \cellcolor{gray!20}25.0 (25.0) & 25.0 (25.0) & 25.0 (25.0) & 25.0 (25.0) & 25.0 (25.0) \\
\midrule
\footnotesize{\textit{\textbf{Any text model}}} \textsuperscript{\dag} & \cellcolor{gray!20}0.0 (0.0) & 0.0 (0.0) & 0.0 (0.0) & 0.0 (0.0) & 0.0 (0.0) \\
\midrule
\multicolumn{6}{c}{\footnotesize \textit{\textbf{Single-Frame Baseline} (access to question, answer choices, and a single key frame from the video.)}}\\
LLaVA-OV \scriptsize{}{(Qwen2-7B)} & \cellcolor{gray!20}11.8 (11.7) & 14.7 (12.2) & 8.7 (10.5) & 2.0 (2.3) & 21.6 (21.9) \\
Qwen2-VL \footnotesize{(7B)} & \cellcolor{gray!20}16.7 (15.7) & 16.9 (13.6) & 20.1 (19.9) & 3.7 (4.4) & 26.3 (24.8) \\
\midrule
\multicolumn{6}{c}{\footnotesize \textit{\textbf{VideoLLMs} (full access to the video, question, and answer choices.)}}\\
LLaVA-OV \scriptsize{(Qwen2-7B)} & \cellcolor{gray!20}20.7 (20.5) & 24.3 (21.8) & 5.2 (5.2) & 5.8 (6.8) & 47.5 (48.2) \\
VideoChat2 \scriptsize{(Mistral-7B)} & \cellcolor{gray!20}23.3 (22.0) & 25.7 (21.0) & 21.4 (20.1) & 10.1 (11.5) & 35.8 (35.3) \\
Mini-CPM-v 2.6 & \cellcolor{gray!20}21.7 (22.3) & 21.3 (20.2) & 18.0 (17.9) & 9.2 (11.9) & 38.3 (39.2) \\
Qwen2-VL \footnotesize{(7B)} & \cellcolor{gray!20}30.0 (29.2) & 27.1 (32.28) & 27.6 (21.2) & 20.0 (18.9) & 45.2 (44.5) \\
LongVU \scriptsize{(LLaMA3-3B)} & \cellcolor{gray!20}20.6 (20.6) & 15.8 (14.1) & 14.8 (16.0) & 16.2 (16.7) & 35.4 (35.8) \\
LongVU \scriptsize{(Qwen2-7B)} & \cellcolor{gray!20}29.9 (29.3) & 28.9 (26.3) & 21.5 (21.8) & 20.5 (22.3) & 48.6 (46.7) \\
Tarsier-7B & \cellcolor{gray!20}26.0 (24.3) & 31.3 (24.5) & 18.7 (18.2) & 15.0 (16.3) & 38.9 (38.2) \\
Tarsier-34B & \cellcolor{gray!20}38.8 (37.4) & \textbf{45.2} (38.7) & 36.3 (36.6) & 21.0 (22.1) & 52.7 (52.4) \\
InternVL2.5-8B & \cellcolor{gray!20}\textbf{40.2} (39.9) & 43.7 (38.1) & \textbf{40.2} (38.7) & \textbf{22.8} (23.1) & \textbf{54.4} (59.8) \\
Gemini-1.5 Pro & \cellcolor{gray!20}--  (29.6) & -- (43.1) & -- (15.5) & --  (19.6) & -- (40.2) \\
GPT4-o & \cellcolor{gray!20} -- (32.5) & -- (36.1) & -- (32.8) & -- (16.2) & -- (45.0) \\
\midrule
Human & \cellcolor{gray!20}92.9 & 91.3 & 91.7 & 97.6 & 90.9 \\
\bottomrule
\end{tabular}
\caption{\textbf{Accuracy on \ourbenchmark{} and \ourbenchmark{}-mini in parentheses}. VideoLLM-performance is slightly greater than random chance, while humans achieve greater than 90\% accuracy on all categories.
Results for closed-source models are only shown on \ourbenchmark{}-mini due to API costs.
Performance is measured via Minimal Pair Score, wherein a model obtains a score iff the prediction for both QA examples of the pair is correct.
\scriptsize{\textsuperscript{\dag} = if temperature of LLM is zero.}}
\label{tab:benchmark_results}
\end{small}
\end{table*}

We evaluate several state-of-the-art open-source VideoLLMs on \ourbenchmark{}, summarized in \Cref{tab:benchmark_results}: LLaVa-OneVision~\citep{li2024llava}, VideoChat2~\citep{li2024mvbench},
Mini-CPM-v 2.6~\citep{yao2024minicpm}, Qwen2-VL~\citep{wang2024qwen2}, Tarsier~\citep{wang2024tarsier} 7B/34B, LongVU~\citep{shen2024longvu}, InternVL2.5-8B~\citep{chen2024expanding}, Gemini-1.5 Pro~\citep{team2024gemini}, and GPT4-o~\citep{achiam2023gpt}.
Most notably these models differ in their generality: The models we evaluate are either generalist models (GPT4-o, Gemini 1.5), specialized for any visual inputs (LLaVa-OneVision, Mini-CPM, Qwen2-VL, InternVL), or specialized primarily for videos (VideoChat2, LongVU).
We also consider two baselines that are fed single-images, LLaVA-OneVision and Qwen2-VL, as they have been trained to process both single image and video.
Note that we additionally evaluate on a smaller balanced version of \ourbenchmark{}, dubbed \ourbenchmark{}-mini, with 1/3 of the original size.\footnote{We release \ourbenchmark{}-mini for fast eval and lower costs of API models.}

\noindent\textbf{Overall performance of VideoLLMs.}
Despite their strong performances on other video QA benchmarks~\citep{li2024mvbench, liu-etal-2024-tempcompass, mangalam2024egoschema, xiao2021next}, \Cref{tab:benchmark_results} shows that most models perform around random chance ($25\%$ accuracy) with the exception of the Tarsier-34B model and InternVL2.5, reaching an average accuracy of $38.1\%$ and $40.2$ respectively.
This is in contrast to human performances which obtain an average accuracy of 92.9\% on a representative subset of \ourbenchmark{} (cf.~\Cref{app:human_annotation}).

While average performance is close to random for most models, we do observe non-trivial performance on several sub-tasks and data sources.
In particular, VideoLLMs achieve better than random performance on \textit{Coarse-grained temporal reasoning}, meaning they possess some ability to distinguish the order of events in a video.

All models fall short on \textit{Fine-grained robot-object interactions}, which involves understanding fine-grained object manipulation on a table with a robotic arm.
This is particularly interesting given the proliferated usage of multi-modal LLMs for learning large-scale visuomotor control policies~\citep{Driess2023PaLMEAE,jiang2023vima}.
Most notably, the \textit{Intuitive physics} category of \ourbenchmark{} is by far the hardest with sub-random scores.
As highlighted by previous works, intuitive physics reasoning is known to be a difficult task~\citep{intphys,grasp,weihs2022benchmarking,du2023videolanguageplanning}, as this involves reasoning about e.g. object permanence, gravity and trajectories.

\noindent\textbf{VideoLLMs performance on dataset sub-tasks.}
Some sources in \ourbenchmark{} are further divided into more fine-grained splits, where each split tests for a specific ability (e.g., object permanence, shape consistency, motion consistency, etc.).
In this section we summarize more detailed observations we gathered on these splits. 

While performance on all intuitive physics tasks is close to 0\%, we find that LongVU (Qwen2) obtains non-trivial performance on three splits:
Gravity-Continuity (39.1\%) and Unchangeableness (42.2\%); with Tarsier-34B performing well on Gravity-Support (35.2\%).
Even some of the weaker models can achieve performance clearly above random on our Fine-grained human-object interactions category when looking closer into subsets such as Counterfactual (e.g., Qwen2-VL: 46.5\% LongVU (Qwen2): 43.9\%) and Memory (e.g., LongVU (Qwen2): 40.4\%) examples.

\noindent\textbf{Importance of Data Curation.}
In \Cref{tab:ablate_filtering}, we explore the effects of the minimal-change pair mining and single-frame bias filtering on model performance.
For this exploration we use the smaller \ourbenchmark{}-mini (see \Cref{app:mini}) and report the average performance of five VideoLLMs \footnote{LLaVA-OV, VideoChat, Qwen2-VL, LongVU (Qwen2), Tarsier-7B}.

When pairing videos randomly instead of using minimal-change pairs, the average accuracy across tasks is at 45.4\%, far superior to random chance.
Using minimal-change pairs, the average VideoLLM performance significantly drops to 27.3\%.
This result shows the importance of the minimal-pair framework and suggests that VideoLLMs can frequently leverage shortcut solutions or spurious features to solve QA tasks.
Additionally, the average VideoLLM performance drops again by another 2.2\% to 25.1\% by removing single-frame solvable videos, with much larger drops on certain subsets.
Note that while \textit{Fine-grained robot-object interactions} and the \textit{Intuitive physics and collisions} categories contain almost no single-frame biases, we can see significant drops of 3.5\% and 3.3\% for the other two categories (\textit{Fine-grained human-object interactions} and \textit{Coarse-grained temporal reasoning}) with this additional filtering step.
Overall, Table~\ref{tab:ablate_filtering} confirms that the minimal-change pair mining and single-frame filtering  pipeline is effective at mitigating potential shortcut solutions in \ourbenchmark{}.

\begin{table*}[h]
\begin{small}
\centering
\setlength{\tabcolsep}{4pt} 
\begin{tabular}{lccccc}
\toprule
\footnotesize \textbf{Model} & \cellcolor{gray!20} \footnotesize \makecell{\textbf{Overall}} & \footnotesize \makecell{\textbf{Fine-grained} \\ \textbf{human-object interactions}} & \footnotesize \makecell{\textbf{Fine-grained} \\ \textbf{robot-object interactions}} & \footnotesize \makecell{\textbf{Intuitive physics} \\ \textbf{and collisions}} & \footnotesize \makecell{\textbf{Coarse-grained} \\ \textbf{Temporal reasoning}} \\
\midrule
\multicolumn{6}{c}{\textit{Pairing of random videos (with same question)}}\\
\small Avg. VideoLLM Acc. & \cellcolor{gray!20}45.4 & 36.8 & 40.9 & 19.7 & 84.3 \\
\midrule
\multicolumn{6}{c}{\textit{+ Pairing of minimally different videos}}\\
\small Avg. VideoLLM Acc. &  \cellcolor{gray!20}\parbox[c]{0.65cm}{\centering 27.3 \\ \scriptsize{\textcolor{red}{\(\downarrow\)18.1}}} & \parbox[c]{0.65cm}{\centering 28.7 \\ \scriptsize{\textcolor{red}{\(\downarrow\)8.1}}} & \parbox[c]{0.65cm}{\centering 18.6 \\ \scriptsize{\textcolor{red}{\(\downarrow\)22.3}}} & \parbox[c]{0.65cm}{\centering 16.7 \\ \scriptsize{\textcolor{red}{\(\downarrow\)3.0}}} & \parbox[c]{0.65cm}{\centering 45.1 \\ \scriptsize{\textcolor{red}{\(\downarrow\)39.2}}} \\
\midrule
\multicolumn{6}{c}{\textit{+ Remove single-frame-solvable examples = final version of \ourbenchmark{}}}\\
\small Avg. VideoLLM Acc. &  \cellcolor{gray!20}\parbox[c]{0.65cm}{\centering 25.1 \\ \scriptsize{\textcolor{red}{\(\downarrow\)2.2}}} & \parbox[c]{0.65cm}{\centering 25.2 \\ \scriptsize{\textcolor{red}{\(\downarrow\)3.5}}} & \parbox[c]{0.65cm}{\centering 18.3 \\ \scriptsize{\textcolor{red}{\(\downarrow\)0.3}}} & \parbox[c]{0.65cm}{\centering 15.2 \\ \scriptsize{\textcolor{red}{\(\downarrow\)1.5}}} & \parbox[c]{0.65cm}{\centering 41.8 \\ \scriptsize{\textcolor{red}{\(\downarrow\)3.3}}} \\
\bottomrule
\end{tabular}
\caption{\textbf{We ablate the effect of our main curation steps.} Both the automatic pairing of minimal pairs and the single-frame-bias filtering lead to lower average model performance, with an especially large drop once we introduce the minimal pair setup.}
\label{tab:ablate_filtering}
\end{small}
\end{table*}

\section{Related Work}

\noindent\textbf{Language biases in Vision-Language models.}
Vision-and-language benchmarks, such as Visual Question Answering (VQA)~\citep{antol2015vqa, goyal2017making, marino2019ok} have been found to be vulnerable to language biases as evidenced by the performance of ``blind" language-only models.
Blind models are routinely shown to be efficient at solving many of the vision-and-language tasks~\citep{goyal2017making, zeng2023socratic, Chen2024-pn}, and can also solve several image-text retrieval benchmarks~\citep{yuksekgonul2022and,hsieh2024sugarcrepe} using language biases~\citep{lin2023revisiting}.
Visual Question Answering in the video-language domain (Video-QA)~\citep{li2024mvbench, he2024mmworld, xiao2021next, lei-etal-2018-tvqa, majumdar2024openeqa, tapaswi2016movieqa,Rawal2024CinePileAL} also exhibits language biases, as shown in the performance of strong language-only baselines~\citep{zhang2023simple, cores2024tvbench}.

\noindent\textbf{Vision-centric biases in Vision-Language models.}
State of the art vision-language models are shown to be surprisingly unaware of the vision inputs, where they often struggle with simple questions due to incorrect visual grounding~\citep{tong2024eyes}, despite leveraging sufficiently powerful visual embeddings.
VLMs are shown to be imprecise at spatial information understanding and geometry~\citep{rahmanzadehgervi2024vision,kamath-etal-2023-whats}.
Similar biases exists in video-and-language tasks, where VideoLLMs typically exhibit single-frame bias~\citep{buch2022revisiting,lei-etal-2023-revealing} or spatial bias~\citep{cores2024tvbench}, where either a single frame is enough to solve the task, or the ordering of the frames is not important. 
To overcome this bias, benchmarks propose computing temporal certificate sets~\citep{mangalam2024egoschema}, key-frame bias~\citep{buch2022revisiting}, or investigate temporal understanding through shuffled frame inputs~\citep{cores2024tvbench}. 
In \ourbenchmark{}, we operationalize a looser definition of temporal understanding for our filtering pipeline (\Cref{sec:benchmark}) in that we keep an example if it is only solvable given the right \textit{key-frame}, but discard it if it can be solved with any randomly sampled frame --- the intuition being that key-frame identification can already involve temporal reasoning.

\noindent\textbf{Benchmarks addressing vision-and-language biases.}
Several approaches are proposed in the literature to reduce the aforementioned biases in Vision-Language systems.
One promising approach is to use minimally different pairs of inputs~\citep{thrush2022winoground, yuksekgonul2022and, hsieh2024sugarcrepe, krojer-etal-2022-image, wang2023equivariant}, also known as Contrast Sets~\cite{gardner-etal-2020-evaluating}, which stem from related work in natural language processing~\citep{levesque2012winograd, sakaguchi2021winogrande, mccoy-etal-2019-right}.
Minimally different input pairs restrict the models' abilities to use these biases, as \textit{both} samples in the pair must be answered correctly to achieve a non-zero score.
Similar to \ourbenchmark{}, some highly adopted examples of such image-language benchmarks build on top of existing image sources (ARO \citep{yuksekgonul2022and}), or fix them explicitly (SugarCREPE \citep{hsieh2024sugarcrepe}).
Commonly, the focus is on \textit{textual} minimal-change pairs, e.g., providing several answer candidates for a question with only slight variations in word order~\citep{yuksekgonul2022and, cores2024tvbench, park2022exposing, li2023vitatecs, cai2024temporalbench}.
However, textual minimal-change pairs can be susceptible to the same language biases~\citep{hsieh2024sugarcrepe, wu2023role}.
Other works, such as in Video-QA, focuses on visual minimal-change pairs.
TempCompass \citep{liu-etal-2024-tempcompass} creates a small set of less than 0.5K artificial minimally different videos by manipulating the original video, e.g., playing the video in reverse, at a faster speed, or playing one video above the other.
Vinoground \citep{zhang2024vinoground} scrapes 0.5K minimally different video pairs from YouTube with the majority following the same pattern: \textit{event A before B} vs.~\textit{event B before A}. 
Our work differs in several aspects from these (summarized \Cref{tab:benchmark_comparison}), notably as well in terms of the scale of curation by showing that minimal video pairs can be procedurally extracted from existing video sources.
While our Minimal Pair Score is inspired by Winoground~\citep{thrush2022winoground}, unlike Vinoground, we intentionally do not adopt the Winoground metric directly since we want \ourbenchmark{} to be agnostic to whether models can process several videos in one forward pass.

The language biases in existing vision-language benchmarks often stem from the over-reliance on world knowledge and plausible co-occurrences~\citep{hsieh2024sugarcrepe, goyal2017making}.
Thus, \ourbenchmark{} focuses on short videos with ``basic'' perceptual skills (spatial, temporal, or intuitive physics), which requires understanding of physical world properties~\citep{Yi2020CLEVRER, chen2021comphy, grasp, intphys, physion, margoni_voe_2024, baillargeon1985object}, reducing the space for blind LLMs to rely on their cultural knowledge.

\section{Discussion and Limitations}
\label{sec:conclusion}

Going back to our initial question, our results suggest that VideoLLMs do not yet perceive and understand the world as reliably as humans.
After evaluating various state-of-the-art VideoLLM models for physical world understanding on \ourbenchmark{}, the best model obtains only 40.2\% average accuracy, while human performance is 92.9\%. 
Yet, VideoLLMs are not completely blind.
On some sub-categories of spatio-temporal understanding and intuitive physics, VideoLLMs can perform significantly better than random chance.
Overall, our empirical evaluation shows that current VideoLLMs are still far from matching human performances on all tested tasks, calling for more research in this direction to develop better training data for world modelling, as well as novel learning criteria and model architectures.
We anticipate \ourbenchmark{} to help the development of the next generation of visual systems to perceive the world as robustly as humans.

\noindent\textbf{Limitations:}
No benchmark comes without limitations.
First, it is possible that forcing the model to output a single letter without any room for free-form reasoning (CoT) limits its performance.
Additionally, using an automated curation approach will not be able to fully remove noisy examples; through manual inspection, we found some of the examples to be too simple, and a few others to be ambiguous, although we note that these noisy samples only represent a small subset of the overall data.

\FloatBarrier
{
    \small
    \bibliographystyle{assets/plainnat}
    \bibliography{main}
}

\newcounter{mypage}
\setcounter{mypage}{\value{page}}

\clearpage
\appendix

\clearpage
\setcounter{page}{1}

\appendix

\section*{Outline}

\textbf{\Cref{app:mini}} explains why and how we created a smaller version of \ourbenchmark{} (\ourbenchmark{}-mini).

\noindent
\textbf{\Cref{app:details_curation}} goes into all the nitty-gritty details of how we curated \ourbenchmark{}.

\noindent
\textbf{\Cref{app:shortcut_details}} contains implementation details and further explorations for the shortcut analysis in \Cref{sec:shortcuts} that motivated building \ourbenchmark{}.

\noindent
\textbf{\Cref{app:human_annotation}} explains how we arrived at the human baseline accuracy on \ourbenchmark{}.

\noindent
\textbf{\Cref{app:inference_details}} provides details on how we prompt VideoLLMs (or single-image VLMs for the shortcut baselines) and extract their answer.

\noindent
\textbf{\Cref{app:behind_scenes} (Behind the Scenes)} shows not just the final product (this paper) but also how we arrived here, what we discarded, and some personal reflections.

\section{\ourbenchmark{}-mini}
\label{app:mini}
Next to the full \ourbenchmark{}, we also release \ourbenchmark{}-mini downsampled in a subset-aware manner to 18,290 video-QA examples (thus 9,145 pairs).
\ourbenchmark{}-mini will be faster to use, while \ourbenchmark{}-full allows researcher to filter and curate derivatives at a large scale.
For the most part we select a random subset, except that we ensure that no dataset (or subset of a dataset) is underrepresented due to subsampling.

\section{Details: The Curation of \ourbenchmark{}}
\label{app:details_curation}
As illustrated in \Cref{sec:benchmark}, we follow three main steps to curate \ourbenchmark{}:
1) Manual categorical filtering
2) Automatic pairing of minimally different examples
3) Automatic filtering of single-frame-solvable examples.
We also need to QAify 5 out of the 9 datasets.

While there are commonalities across datasets for how we implement Step 1 (categorical filtering) and Step 2 (pairing), there is also some differences that we describe here for reproducibility.
Note that Step 3 is exactly the same across datasets, and is described in sufficient detail in \Cref{sec:benchmark}.
Hence, we focus on the first and second step.

\subsection{Manual categorical filtering}
\label{app:manual_filter}
For 6 out of 9 datasets we select subsets and categories of questions suitable for \ourbenchmark{}.

\textbf{Perception Test:} We manually annotate the 132 question types in Perception. Specifically we filter out question that either do not require temporal understanding (``Where is the person?'') or are ill-defined. Around 20\% are discarded.

\textbf{Language Table:} We select the human-captioned and human-controlled split of Language Table which constitutes 440K. Thus we exclude other splits where the robot arm is automatically controlled and/or the robotic actions are synthetically captioned and not by a human.
Additionally, we exclude any videos where the caption only mentions a single object such as ``Move the arm to the left'' to ensure complex enough interactions.

\textbf{CLEVRER: } We find an issue in the counterfactual split of CLEVRER and exclude these examples based on the meta-data associated with each video, e.g. object attributes and the exact position of each object at each frame.
The issue is that most of the time, the object mentioned (\textit{target object}) in the question (``What happens if the cube is not there?'') is not actually involved in any collisions.
As a result, the correct answer (e.g. ``The red and yellow cube collide'') is often depicted, whereas in a proper counterfactual example the correct answer should never be depicted but only happen in an ``alternative world''.
Thus we filter out an example if the target object is never near any other moving objects (i.e. a collision) based on their coordinates.
Moreoever, we filter out an example if the two objects mentioned in the correct answer are in fact already colliding in the video, based on their coordinates.

\textbf{InfLevel: } We only use the split with real-life videos where humans conduct the experiments in front of a camera, similar to experimental designs in psychology \citep{weihs2022benchmarking}.

\textbf{STAR: } We exclude the \textit{Feasibility} and \textit{Interaction} splits since they are often ill-defined, lead to strong language biases or are too easy.

\subsection{Automatic Pairing of Minimally Different Examples}
\label{app:minimal_filtering}

We apply this step to only 5 out of the 9 datasets, since Vinoground \citep{zhang2024vinoground} and the 3 intuitive physics datasets are already structured into minimal-change video pairs.
Our pairing boils down to finding highly similar pairs, which we base on symbolic or visual similarity, and at the same time ensuring that both answers cannot be true at the same time, i.e. \textit{mutually exclusive}. Especially the latter task, also known as \textit{entailment} detection or \textit{natural language inference}, has many nuanced edge cases.

This step is conducted on QA examples $x$ consisting of a question, a video, and answer candidates: \((q, v, [\text{a}_1, \text{a}_2, \ldots])\).

\noindent
\paragraph{Perception Test/STAR.}
\begin{enumerate}
    \item Group QA examples into sets with the same question:
    \[
    P = \{ X \mid \forall x_i, x_j \in X, q_i \equiv q_j \}.
    \]

    \item For a given \(X\), examples \(x_i\) and \(x_j\) are grouped into potential pairs if they have opposite (\textit{mutually exclusive}) correct answers:
    \[
    P' = \{ (x_i, x_j) \mid a_i \neq a_j \}.
    \]

    \item From this set of potential pairs \(P'\), we choose the top-\(k\) for a given question based on visual similarity, measured via cosine similarity of embeddings from the video encoder ViCLIP-ViT-L~\citep{wang2023internvid}:
    \[
    \begin{aligned}
    P_k = \big\{ & (x_{i_1}, x_{j_1}), \ldots, (x_{i_k}, x_{j_k}) \;\big|\; \\
    & \text{sim}(v_{i_m}, v_{j_m}) \geq \text{sim}(v_{i_{m+1}}, v_{j_{m+1}}) \big\}.
    \end{aligned}
    \]
    In practice, we choose \(k=50\) for each question.
    
\end{enumerate}
Additionally, we use dataset-specific rules after manual inspection, e.g., for the Perception Test, we require that for two potential pairs \(x_i\) and \(x_j\), neither correct answer \(a_i\) nor \(a_j\) is ``Both the other options".

\noindent
\paragraph{Language Table.}
Note that all examples in Language Table have the same question ``Which robot instruction best describes the actions in the video?''.
\begin{enumerate}
    \item We group QA examples into sets such that a) both correct answers mention the same objects (e.g., both involve a ``red triangle'' and ``green heart'') and b) the set of tokens in \(a_i\) and \(a_j\) have a large enough overlap: 
    \[
    \begin{aligned}
    P = \big\{ & (a_i, a_j) \;\big|\; 
    \text{obj}(a_i) \equiv \text{obj}(a_j) \land \\
    & 0 < \text{token\_diff}(a_i, a_j) < 4 \big\}.
    \end{aligned}
    \]
    Due to the finite number of attributes and objects in Language Table, \(\text{obj}(\cdot)\) checks for these attribute and object key-words.

    \item We narrow this set of potential pairs \(P\) with a visual similarity threshold, measured via cosine similarity of embeddings from the video encoder ViCLIP-ViT-L~\citep{wang2023internvid}: 
    \[
    P' = \left\{ (x_i, x_j) \mid \text{sim}(v_i, v_j) > 0.9 \right\}.
    \]

    \item Finally, we ensure that answers are mutually exclusive, i.e., \(a_i \not\implies a_j\) and \(a_j \not\implies a_i\). In practice, this involves several hand-crafted rules after inspecting failure cases: If the order of objects mentioned is different, there is no entailment (e.g., ``Move yellow triangle to blue heart'' and ``Move blue heart to yellow triangle''); if otherwise one answer contains a \textit{general direction} such as ``towards'', ``to'' or ``into'' but the other answer contains a \textit{specific direction} such as ``left'' or ``above'', there is entailment (we discard the example). To illustrate: ``move the X towards Y'' entails ``move X to the left of Y''. We identify several of such situations.
\end{enumerate}

\noindent
\paragraph{Something Something v2.}
Note that Something Something v2 is a video caption dataset where each caption contains either one or two objects and a simple action, with in total 174 types of such actions. We QAify these examples with the question ``Which action is being performed in the video?`` and use the caption with something-placeholders instead of objects as the answer $a$, e.g., ``dropping something''.

\begin{enumerate}
    \item We group QA examples $a_i$ and $a_j$ into pairs such that the action in $a_i$ is a well-defined antonym of the action in $a_j$: 
    \[
    P = \{ (a_i, a_j) \mid \text{antonym}(a_i, a_j) \}.
    \]
    In practice, we identify a subset of 82 action types (47\% of all actions) that have a well-defined opposite, e.g., ``spinning something so it continues spinning'' and ``spinning something that quickly stops spinning''.
    
    \item We narrow down pairs further by selecting a pair $x_i$ and $x_j$ if the videos contain the same object(s) based on their captions:
    \[
    P' = \{ (x_i, x_j) \mid \text{obj}(v_i) \equiv \text{obj}(v_j) \}.
    \]
    If no pairs fulfill this strict criterion, we relax it such that only one object must overlap:
    \[
    P' = \{ (x_i, x_j) \mid \text{obj}(v_i) \cap \text{obj}(v_j) \neq \emptyset \}.
    \]

    \item From this set of potential pairs \(P'\), we choose the top-\(k\) based on visual similarity, measured via cosine similarity of embeddings from the video encoder ViCLIP-ViT-L~\citep{wang2023internvid}:
    \[
    \begin{aligned}
    P_k = \big\{ & (x_{i_1}, x_{j_1}), \ldots, (x_{i_k}, x_{j_k}) \;\big|\; \\
    & \text{sim}(v_{i_m}, v_{j_m}) \geq \text{sim}(v_{i_{m+1}}, v_{j_{m+1}}) \big\}.
    \end{aligned}
    \]
    In practice, we choose \(k=4000\).
\end{enumerate}

\noindent
\paragraph{CLEVRER.}
Note that CLEVRER has detailed meta-data with a list of all objects throughout the video and their attributes (color, shape, material), with many videos featuring five or more objects.

\begin{enumerate}
    \item Group QA examples into sets with the same question:
    \[
    P = \{ X \mid \forall x_i, x_j \in X, q_i \equiv q_j \}.
    \]

    \item For a given \(X\), examples \(x_i\) and \(x_j\) are grouped into potential pairs if they have opposite (\textit{mutually exclusive}) correct answers:
    \[
    P' = \{ (x_i, x_j) \mid a_i \neq a_j \}.
    \]
    In the special case that the answers are both numerical, we require them both to be 1 apart, e.g., ``How many objects are moving when the video ends? A) 2 B) 3''.

    \item We further filter the set \(P'\) by requiring a large overlap of objects with the exact same attributes in both videos. Specifically, we keep a pair if the set of objects in $v_i$ is a ``fuzzy subset'' of the objects in $v_j$, or vice versa:
    \[
    \begin{aligned}
    P'' = \{ (x_i, x_j) \mid \text{fuzzy\_subset}(\text{obj}(v_i), \text{obj}(v_j)) \lor  \\
    \text{fuzzy\_subset}(\text{obj}(v_j), \text{obj}(v_i)) \}.
    \end{aligned}
    \]
    Here, $\text{fuzzy\_subset}(\cdot, \cdot)$ allows one mismatch between the sets of objects and their attributes.
\end{enumerate}

\section{Details: Shortcut Analysis}
\label{app:shortcut_details}

We provide additional experiments for the shortcut analysis on MVBench datasets from \Cref{sec:shortcuts}.

\paragraph{Language only shortcuts.} Similar to \citep{li2024mvbench} we also tested a VideoLLM as a language-only baseline by blacking out the video (replacing it with zeros). With VideoChat2 this gave slightly worse results with both LLM-versions: 33.0\% (Mistral) and 34.6 (Vicuna).

\paragraph{Video only shortcuts.}
For the video only we remove the question and only provide the answer candidates to the model.
In detail, we tested several ways of removing the question (empty string, replace with ``what?'', explain to answer without question, etc) and found that replacing the question with ``[REDACTED]'' yielded the best performance.

\paragraph{Single-frame only shortcuts.}
First we study how performance varies when selecting frames at different positions: first, middle, last, random and finally key-frame. We choose a key-frame based on the highest CLIP similarity among all frames and all answer candidates (with the question prefixed to the candidate).
We find that the performance differs only by 1-2\% among these selection strategies except for the first frame which performed more than 5\% worse than the rest with Idefics3.
Since middle has the highest MVBench accuracy, excluding key-frame, we show middle frame results in the paper.
In the main paper we show results for Idefics3 and Qwen2-VL, models that mostly focus on non-video tasks.
We also test VideoChat2 variants but found performance to be worse, with either showing a single frame once or copying it 16-times as a ``video''.

\paragraph{Simple Socratic LLM shortcuts.}
In this shortcut we test how well models can still perform when the video is replaced by a much lower bandwidth representation and presented to a text-only LLM: a short generic caption of the video.
We generate these captions with InternLM-XComposer-2.5-7B~\citep{zhang2024internlm}.
We investigate how model performance differs when increasing the bandwidth of this caption from short, medium to long caption:
``[...] Briefly describe this video in one sentence.'', ``[...] Describe this video in 1-2 sentences.'' and ``[...] Describe this video in as much detail and length as possible.''.
We also ask whether focus on objects or actions helps, i.e. by prompting the captioning model to list the objects or actions in the video.
While the long caption variant achieves the highest performance when provided to LLaMA3 8B and 70B, followed by action caption, we choose to show the medium caption variant (i.e. asking the model to caption the video in 1-2 sentences) in the main paper since this is most in the spirit of a short (1-2 sentences) and generic (not asking for anything specific) caption as a simple baseline.

\paragraph{Additional robustness experiments beyond main paper.}
As a sanity check, we also study how well a perplexity baseline and answer frequencies perform on MVBench.
For the perplexity baseline we compute the perplexity of each answer candidate sequence based on LLM (LLaMA3-7B), i.e. how plausible this string is by itself.
For example, are common objects or scenarios more often the correct answer? This would be reflected in such a baseline.
However we find that this baseline performs around random overall.
Next, we also compute statistics to determine if some subtasks of MVBench have a skewed distribution of answer frequencies, i.e. whether option A is more often correct than the other option B, C, etc. or if it is more often ``yes'' than ``no'', etc.
Here we also also find very little evidence of any issues in terms of frequencies.


\section{Human Annotation}
\label{app:human_annotation}
We assigned the videos to 6 researchers from our lab, recorded their answer responses, and then computed the benchmark metric using the pair-wise scoring.
Each person was assigned one video from the pair at a time, thus avoiding any advantage over VideoLLMs that would come from seeing both videos in a minimal-change pair (i.e., avoiding any knowledge that the answer can only be AB or BA).

\clearpage
\section{Details on prompting multi-modal LLMs}
\label{app:inference_details}
For evaluating VideoLLMs, we use the following prompt (with an example question from our benchmark):

\begin{tcolorbox}[colback=white, colframe=black, title=VideoLLM Prompt]
You are an expert video understanding AI system. Carefully watch the video and pay attention to the cause and sequence of events, the details and movements of objects, and actions of people. Based on your observations, select the best option that accurately addresses the following question: 

Q: \texttt{\{Question\}}\\
A) \texttt{\{Correct answer for video1\}}\\
B) \texttt{\{Correct answer for video2\}}

Even when unsure, always answer with a single letter from A or B, format exactly like: `Answer: A/B'.
\end{tcolorbox}

We extract the answer letter via a simple regex and find that this approach fails in only less than 1\% of examples.

\section{Behind the Scenes}
\label{app:behind_scenes}

In this section we go beyond what usually goes into a paper and discuss how the paper came about, what did not work, or what motivated the authors - so in essence: all the things that are usually deemed too subjective or ``unscientific'', yet would help other researchers, especially those joining the field, often much more than the polished narrative of the main paper.

\subsection{Motivation and timeline}
Several of the authors who work on video modeling felt a growing frustration with existing benchmarks that often rewarded the wrong things.
So the direction of the project was quickly set after a short period of brainstorming in June and July:
Quantify in what ways existing benchmarks are broken, and then fix it.
We spent July - September staring at hundreds of examples from the MVBench datasets, scouting for glaring issues or shortcuts and manually annotating lots of data.
First, we tested the simple baselines wrt. frequency, text-only or single-frame biases, and soon included the less often discussed video-only (remove question) and Simple Socratic LLM shortcuts.
In between we had philosophical discussions about benchmark design (bottom-down vs. top-down) or what it means for a video task to be truly temporal: is it temporal if two-frames are needed, or if a single frame is needed but it has to be a key-frame (needle in a haystack), or ...?
Regarding benchmark design, should we adopt other paper's taxonomies or design or own? Should one collect all kinds of examples and ad-hoc define a taxonomy (bottom-up), or should one define a taxonomy, then systematically collect examples to fit the taxonomy (top-down)?
From the beginning the idea of minimal video pairs generated excitement among us:
Minimal visual pairs have led to much progress in the field of vision-and-language compositionality (e.g. Winoground), yet had not been explored much in the realm of moving images.

There were, and perhaps still are, plans to crowd-source human shortcut performance, i.e. how much better are humans at solving video tasks when given single frames? At scale this could also be used to filter out examples with more precision than our five-model ensemble approach.
Doing human crowd-sourcing well is not trivial, it is time-intensive and requires dedication but it can lead to much stronger insights than relying purely on automatic metrics and black-box models.

After the exploration phase, we executed on the benchmark building from September to November:
From the start we had identified several promising datasets to mine minimal video pairs and continuously added a new source roughly every week. Perception Test was the first to go through our curation pipeline and hence took the longest as we were still refining the pipeline steps. Language Table was very hard to do well with many edge cases in the entailment detection, and also with its scale of 440K video-caption examples (imagine looking for potential pairs of videos, i.e.  $440,000^2$ combinations).

\subsection{Observations and lessons learned}
\begin{enumerate}
    \item There is too many edge cases to catch every single one at this scale of data curation.
    \item Paper writing is smooth when the story and contribution is clear from the beginning of the project (This was not the case in the first author's last paper so it was nice to observe the contrast).
    \item The intuitive physics datasets are (to the subjective taste of the first author) the cleanest and most fascinating sources in \ourbenchmark{}.
    \item At the same time, Something Something v2 truly stood the test of time as a great video understanding dataset due to  its low noise-ratio at a scale of 200K examples, and coverage of interesting yet simple phenomena.
    \item Frame rate plays a big role for solving many examples in \ourbenchmark{}. In order to push the field further we are now asking the models more and more nuanced questions, and the answer may lie only in a short span of a less than second. However many models may not have access to this short span in principle as they represent a video as 16 uniformly sampled frames.
    \item Parsing outputs from LLMs into a structured format such as answer options can feel like the wild west sometimes: Could models perform better if we prompt them better, or have more flexible ways of extracting the answer?
\end{enumerate}

\subsection{Advice for others working on a similar direction}
Video-QA is becoming a crowded field with several new benchmarks every month. So think twice before you start a project! More seriously, it is a very exciting direction with enough enough dimensions for everyone to innovate on: long video benchmarking, intuitive physics, social commonsense/Theory-of-Mind/narratives, novel simulation engines, ...

Despite our best efforts studying shortcuts, we probably missed some shortcuts or issues in \ourbenchmark{}. It is good to think two steps ahead what kind of shortcuts future more capable models could take. It is easier said than done, but in retrospect older video benchmarks from 2015-2020 might have been able to address single-frame biases a priori during benchmark design.
Instead now the field took years to identify and clean up benchmarks.

\FloatBarrier

\end{document}